# Newsvendor Model with Deep Reinforcement Learning

Dylan Goetting

Dylangoetting@berkeley.edu

**Abstract**

I present a deep reinforcement learning (RL) solution to the mathematical problem known as the Newsvendor model, which seeks to optimize profit given a probabilistic demand distribution. To reflect a more realistic and complex situation, the demand distribution can change for different days of the week, thus changing the optimum behavior. I used a Twin-Delayed Deep Deterministic Policy Gradient agent (written as completely original code) with both an actor and critic network to solve this problem. The agent was able to learn optimal behavior consistent with the analytical solution of the problem, and could identify separate probability distributions for different days of the week and behave accordingly.

## 1. Introduction

The Newsvendor model is a classic problem in operations research and applied economics, in which the demand for a product is defined with a probability distribution, and the goal is to determine the optimal inventory level to stock. This depends on the distribution, the production cost of the product, and the profit margin of the product. Analytically, this problem can be solved with a simple formula, but this assumes knowledge of a perfect probability distribution, which isn't the case in the real world. Thus, a data-driven approach is more practical and should have better, more realistic results.

Several supervised deep learning models have been proposed - Lu & Chang (2014), Ban & Rudin (2017) - in which a model is trained to forecast the demand, but these solutions ignore a key aspect of the problem [4]. Depending on the relative magnitudes of the production costs and profit margins, the optimal inventory level should be higher or lower than the expected demand. Additionally, the standard deviation of the distribution is an important factor, as distributions with a higher standard deviation should push the optimal level farther away from the mean. For example, if the standard deviation is 0, the actual demand will always equal the mean and thus the optimal level is also the mean.

This is why a RL model is valuable, as the agent can observe reward patterns in real time and actually learn what the optimal value is. Further benefits of using this model are that the demand distribution does not need to be known at all, and the agent will be able to adapt to any data it sees regardless of the data source. Simply guessing a parameterized distribution from data can be difficult and can easily create sub-optimal results, which is mostly what is done in practice. On the other hand, an RL model does not need to guess a distribution - it internally learns the exact properties of the data it sees, which is much more reliable.

This paper demonstrates that a RL agent is capable of learning demand distributions, and is able to use the reward signals to achieve optimal behavior. This is the first published RL solution to the problem, and shows that the results from using RL are consistent with the analytical solution.

## 2. Background

### 2.1 Newsvendor

The newsvendor problem is parameterized by *D,* the demand distribution, *p, the retail price, and c, the unit cost.* Thus the value for profit that we seek to maximize is calculated by

$$\mathrm{E}[\text{profit}] = \mathrm{E}[p \min(q, D)] - cq$$

[3]

Where E is the expected value operator and q is the independent variable. The analytical solution to the problem is as follows:

**Critical fractile formula**

$$q = F^{-1}\left(\frac{p-c}{p}\right)$$

[3]

Here, $q$ is the optimal quantity to produce, $F$ is the cumulative density function of the demand distribution, and (p-c)/p is called the *critical fractile.* Thus we want to find a quantity q such that the cumulative density of the demand is equal to the critical fractile. For some distributions, such as exponential and uniform, we can easily find a formula for $F^{-1}$ and quickly solve for q. In the most common case, some form of a normal distribution, it is impossible to derive an expression for $F^{-1}$, but we can use Z-score tables and such methods to quickly obtain the value of q. Still, this approach to the problem relies on the explicit knowledge of the demand distribution beforehand, with certainty that the demand behaves exactly according to that distribution. In any realistic situation, even with large amounts of data available, the exact distribution of the demand will be impossible to determine.

### 2.2 Deep RL

The field of RL is centered around the idea of an *agent* and an *environment,* and the interaction between the two. The environment gives the agent some information, (the state) then the agent uses that information to decide upon an action to take. When the agent chooses that action, the environment then returns a successive state, and potentially some reward. This process is known as a Markov Decision Process (MDP), where an agent moves between different states and collects rewards along the way [1]. It is the agent's job to choose actions that maximize its total reward over the whole process, and there are many ways to do this.

A central concept in RL is the idea of a transition buffer, where the agent keeps track of all the transitions it has seen from interacting with the environment. A transition is basically a tuple of *(state, action, reward, next state)* [1]. This gives the agent a memory of everything it has seen, and uses this to generalize and learn which states and actions are more likely to lead to higher rewards.

Perhaps the most important function is the Q-value function, which is defined as the total cumulative reward that the agent should receive (assuming it behaves optimally), starting from an initial state *s* and taking an initial action *a*. The agent then uses this function to make decisions by looking at the Q values for different possible actions. The Q function obeys a very important property known as the Bellman equation, and can be written as follows:

$$Q^*(s, a) = r(s, a) + \gamma \max_{a'} Q^*(s', a')$$

[8]

Q* represents the expected total reward, r(s, a) is the actual reward the agent receives from taking action a in state s, Q*(s', a') refers to the total reward the agent should receive from the succeeding state, optimized over all possible actions a'. γ is the *discount factor,* a number between 0 and 1 that measures how much to prioritize current versus future rewards [9]. The intuition with this equation is that the total reward from the current state should equal the actual reward experienced, plus the discounted total reward from the succeeding state. This allows us to solve the problem in a recursive way, often using dynamic programming.

In deep RL, the agent uses neural networks to generalize and approximate its functions, and trains from the real-time data it is observing. Thus when the agent needs to evaluate a given state and action pair, it inputs the pair into a neural network that outputs a number: the expected discounted reward from starting in that state and taking that action. It can then use the actual reward and the network's estimate of the total reward from the future state to update the network's parameters using gradient descent.

This method is considered *off-policy,* meaning that the agent can use data from any point in its memory to learn, and not just the current data it is seeing. This makes sense, because the Bellman equation still holds for transitions throughout any point in the agent's history. This is very data efficient, and the agent can sample many different transitions from its replay memory to update the networks, all at once. The same transition can be used over and over again, each time providing good information as to how to improve the network approximation. With this RL methodology, we can begin to train a model to optimize its profit within a Newsvendor setting.

## 3. Problem Setup

### 3.1 Environment
To set up the problem, a simulated demand environment is needed. In RL, the environment must do the following: display some form of 'state' to the agent, internally transition the agent between states, and give the agent feedback (reward) at every step. I constructed an environment object, a simple class which has primary methods *observe*, *reset* and *step*.
- Observe() simply returns the state, which is the primary way in which the agent makes decisions.
- Reset() internally sets the agent's state back to the initial state, which is used at the beginning of every episode.
- Step() is the method in which the agent actually interacts with the environment, moving forward in the simulation. The agent passes in an action, and then the environment runs internal calculations to determine the reward and the succeeding state. Importantly, the agent is blind to the reward structure the agent uses, and the state transition dynamics that determine the next state. Step() returns the reward, next state and a 'done' boolean, and the agent can then use this information to calculate another action, and repeat the process until the episode is done.

### 3.2 Reward Structure
In this specific environment, the reward signal is a measurement of profit, and how well the agent predicted the demand for a given day. It is calculated based on the following: the action chosen, the randomly sampled demand, the unit profit and the unit production cost. The unit costs are both external values that parameterize the environment, represented as instance fields, that depend on the product being modeled. The action passed in is a real number representing the units of inventory to stock. The demand is the interesting part. As part of the environment object, there is a list of distribution objects - a separate class used to represent a probability distribution. A distribution object can really be anything as long as it has a *sample()* method, which generates some sort of demand value. In the objects I have created, the distribution objects import Scipy distributions - i.e. gamma, uniform, normal, etc, and are parameterized by their mean and standard deviation. The environment will then sample from a specific distribution - depending on the state - and that will be the demand for that day. The reward is then calculated from the equation: $profit = p * min(d, a) - c * min(c, a)$. Note this is slightly different from the equation in Figure 1, here *p* represents the profit of selling a unit instead of the selling price.

### 3.3 State representation
The definition of the *state* in an MDP is crucial in determining the behavior of the agent, and engineering optimal behavior. The concept of a good state is to tell the agent meaningful information that should influence the agent's decision. In many RL situations, the state-space can get very large (the game of Go has around $10^{170}$ different states [7]), but in the Newsvendor model it is much simpler.

The state information I chose to incorporate into the model was the 'day of the week', which in the real world is a significant factor that can change the demand distribution dramatically. Importantly, telling the agent the day of the week does not reveal anything about the actual demand distribution or the way in which reward is calculated, instead it gives the agent an indicator value, from which the agent can begin to learn. This is not standard in the original Newsvendor model, but is a modified version of the problem where each day yields a unique demand distribution. This was a conscious decision to make the model more applicable to the real world, where the demand on a Saturday is likely much different than the demand on a Wednesday.

To accommodate this, the environment class contains a seven-element list of distribution objects, which can be externally set to anything. Each timestep, the environment iterates to the next 'day' and then selects the distribution according to the day. The state the agent actually receives is simply a number from 0 to 6, that progresses every timestep. To make this more friendly to the neural networks, the agent then internally converts this to a binary tensor (ie [0, 0, 0, 0, 1, 0, 0]) before propagating it through the networks. In general, networks are able to learn much better when there are more features in the input, and a binary tensor makes it easier to differentiate between the different states.

### 3.4 The algorithm
The actual algorithm I used to solve the problem is Twin-Delayed 3 (TD3), a modification of the better known Deep Deterministic Policy Gradient (DDPG) algorithm.

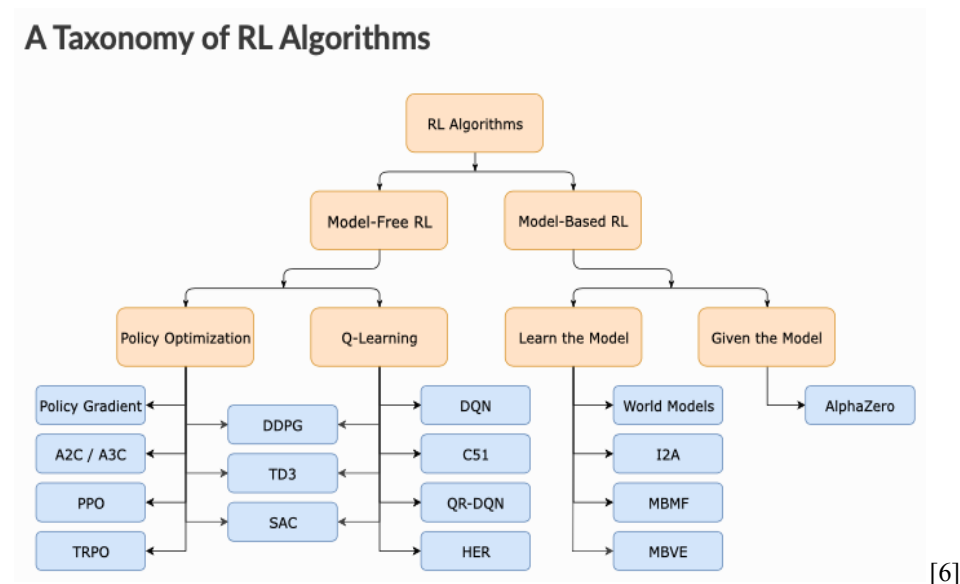

[6]

Figure 1: The taxonomy of most RL algorithms

As seen in the graph, TD3 implements both a Q-learning approach and a Policy optimization approach, meaning it trains two separate neural networks, one for selecting an action, and one for evaluating state-action pairs. They are named the actor and critic networks, respectively. The actor network takes in a 1x7 state tensor, and directly outputs a 0 dimensional number representing the action. The critic network takes in a 1x8 tensor (a 1x7 state concatenated to a 1x1 action) and outputs a 0 dimensional estimate of the Q-value. In order to train, both networks work together and build off each other as described below.

To train the actor network, the sampled output of the actor is directly input into the critic network, and then gradient descent optimizes the parameters to output a more desirable action - one with a higher Q-value. This clearly relies on the critic network to be able to accurately determine the Q-value of a state-action pair, which is itself trained separately. To train the critic network, we look at the real reward observed, and then add that to what the critic estimates the Q-value of the next state. This is compared to the critic's actual prediction, and then gradient descent is used to optimize the parameters to match the target.

DDPG and TD3 both use target networks in the training to avoid divergence. Target networks are older versions of the model that are used when training, which ensures that the networks will be able to converge to optimal behavior. The way target networks are updated is through a *soft update* aka polyak averaging, which looks like this:

$$\phi_{\text{targ}} \leftarrow \rho \phi_{\text{targ}} + (1 - \rho) \phi$$

The parameters of the target network are a weighted average of the previous target parameters and the current parameters, where *p* is usually around 0.005 [5].

TD3 deviates from DDPG in three main ways. The first is the 'twin' aspect, in which the agent actually learns two identical critic networks. When calculating the target Q-values, it uses the smaller of the two. This avoids a common problem in DDPG where the critic is prone to overestimation and the resulting policy diverges [2]. The second is the 'delayed' aspect, where the actor network is updated less frequently than the critic network, usually with a 1:2 ratio [2]. The third is target policy smoothing, where noise is added when calculating the target Q-values, which has the effect of smoothing the output of the critic, making it harder for the agent to exploit noisy errors.

To solve the well known exploration vs exploitation dilemma, I have broken training into three different phases. The first phase is a buffer priming period, where the agent behaves *completely randomly*, solely to generate experience to store in the replay buffer. This explores the vast majority of the action space, and gives the buffer a copious amount of data to work with. The second phase is the training period, where the functions that actually train the networks are called, and the agent begins to learn. The agent now selects actions according to its actor network, and a small amount of noise is added to maintain some exploration. The third phase is the evaluation period, and the agent no longer trains its networks nor adds noise to the action. This provides a very quick way to directly evaluate the performance.

The reason for selecting TD3 was the benefit of both Q-learning and policy optimization. In this environment, the action space is semi-continuous, and when the possible demand range gets large, it is impractical to use a purely value based method. At the same time, Q-learning is still very important, and learning the values of different state action pairs makes it much easier for the agent to learn the demand distributions. Having both an actor and critic network, DDPG gets the best of both worlds, and the TD3 modifications have been proven to improve the performance of DDPG.

Here is general pseudocode for the whole algorithm

```
Algorithm 1 Twin Delayed DDPG
1:  Input: initial policy parameters θ, Q-function parameters φ₁, φ₂, empty replay buffer 𝒟
2:  Set target parameters equal to main parameters θ_targ ← θ, φ_targ,1 ← φ₁, φ_targ,2 ← φ₂
3:  repeat
4:      Observe state s and select action a = clip(μ_θ(s) + ε, a_Low, a_High), where ε ~ 𝒩
5:      Execute a in the environment
6:      Observe next state s', reward r, and done signal d to indicate whether s' is terminal
7:      Store (s, a, r, s', d) in replay buffer 𝒟
8:      If s' is terminal, reset environment state.
9:      if it's time to update then
10:         for j in range(however many updates) do
11:             Randomly sample a batch of transitions, B = {(s, a, r, s', d)} from 𝒟
12:             Compute target actions
```

$$a'(s') = \text{clip}\left(\mu_{\theta_{\text{targ}}}(s') + \text{clip}(\epsilon, -c, c), a_{Low}, a_{High}\right), \quad \epsilon \sim \mathcal{N}(0, \sigma)$$

```
13:             Compute targets
```

$$y(r, s', d) = r + \gamma(1 - d) \min_{i=1,2} Q_{\phi_{\text{targ},i}}(s', a'(s'))$$

```
14:             Update Q-functions by one step of gradient descent using
```

$$\nabla_{\phi_i} \frac{1}{|B|} \sum_{(s,a,r,s',d) \in B} (Q_{\phi_i}(s, a) - y(r, s', d))^2 \quad \text{for } i = 1, 2$$

```
15:             if j mod policy_delay = 0 then
16:                 Update policy by one step of gradient ascent using
```

$$\nabla_\theta \frac{1}{|B|} \sum_{s \in B} Q_{\phi_1}(s, \mu_\theta(s))$$

```
17:                 Update target networks with
```

$$\phi_{\text{targ},i} \leftarrow \rho\phi_{\text{targ},i} + (1 - \rho)\phi_i \quad \text{for } i = 1, 2$$
$$\theta_{\text{targ}} \leftarrow \rho\theta_{\text{targ}} + (1 - \rho)\theta$$

```
18:             end if
19:         end for
20:     end if
21: until convergence
```
[5]

Figure 2: Pseudocode for the TD3 algorithm

## 4. Experiments

**4.1 Analytical Solution**

The first experiment to run is to compare the agent's behavior to the mathematically optimal behavior. In this simpler case, the environment is initialized with seven identical distributions, chosen to be a normal distribution with μ = 50, σ = 20. The unit profit and unit costs are initialized to $2 and $5 respectively. Intuitively, the production cost is greater than the profit, so when dealing with uncertainty it would make sense to be conservative and produce *less* than the mean of 50. To solve this problem analytically, we refer to the critical fractile formula, and solve the Z-score for this distribution:

$$q_{\text{opt}} = F^{-1}\left(\frac{7-5}{7}\right) = \mu + \sigma Z^{-1}(0.285) = 50 + 20(-0.56595) = 38.68 \approx 39.$$

[3]

Now, when running the agent on this distribution we expect the agent to ultimately learn this value, and choose an action of ~39 every time. Since all seven days have the same distribution, there should be no variance in the behavior across the different days. Two key metrics are tracked: the success (profit) of the agent over the course of its training, and the actions it chose for each state. Both graphs are as follows:

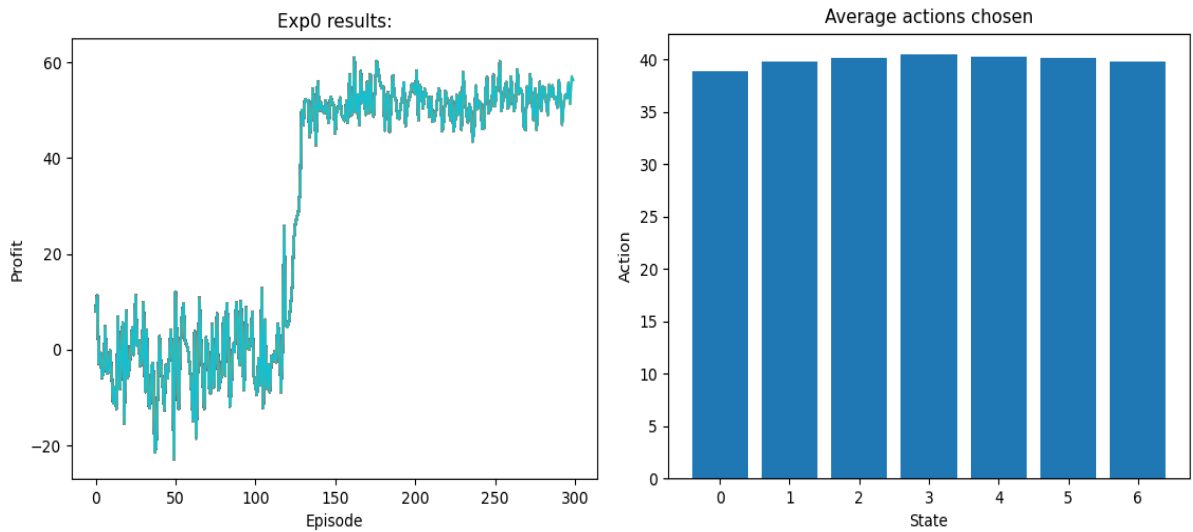

Figure 3: The plots from the results of the first experiment.

The learning curve of the agent (left) illustrates the different phases in the learning process, and the clear progress it makes. In the first segment, the agent is acting randomly, and the reward during this period is mostly negative, and noisy. At around episode 120, the buffer priming period ends, and the agent starts training its networks. Due to the large amount of data in the buffer, the agent is able to learn very quickly and converge upon a high level of profit/reward. From there until the end of the simulation, it remains stable but noisy, due to the randomness of the demand distribution.

The plot on the right shows the agent's behavior during the evaluation phase. It upholds our two main expectations from the analytical solution. First, the actions chosen for each state are all very similar. A small amount of variation is expected, because the randomness in the demand does not guarantee data for each state to be exactly the same. Secondly, the values for the actions are slightly under 40, compared to the mathematical optimum of 39. Out of all the possible actions the agent can take (this experiment restricted it from (0,100)), this provides strong evidence that the agent is able to truly learn optimal behavior from the data it is given. When sampling from a distribution like this, there will always be noise and the data will never fully match the distribution.

**4.2 Different Days**

The second experiment I ran was an aggregate test that built off the previous experiment, with the goal of testing the agent's ability to adapt to different distributions. The seven distributions are set up as follows: Days 0-2 are normal distributions with mean 50 and standard deviations 5, 10 and 20 respectively. These serve to test an important intuition: higher standard deviations should push the optimal level farther away from the mean. If the unit profit exceeds the unit cost, these optimums should be greater than the mean, and vice-versa. Day 3 is an exponential distribution with mean/standard deviation of 5, and shifted over by 30. This tests the agent's ability to learn a different kind of distribution, one that is non-symmetric and very different from a normal distribution. Days 4-6 are a sanity test; these are uniform distributions with means of 70, 80, 90 respectively, and all with standard deviation of 0. Since the demand will be exactly the same every time, the evident solution to this is to choose exactly the mean.

The environment is set up slightly differently this time, with a unit profit of 10 and unit cost of 5. This time, since the profit exceeds the cost, the optimal level should be greater than the mean because correct overestimates will outweigh correct underestimates. The hyperparameters are the same as the first experiment, to keep the model consistent and to avoid overfitting the experiment. Both graphs for this experiment are as follows:

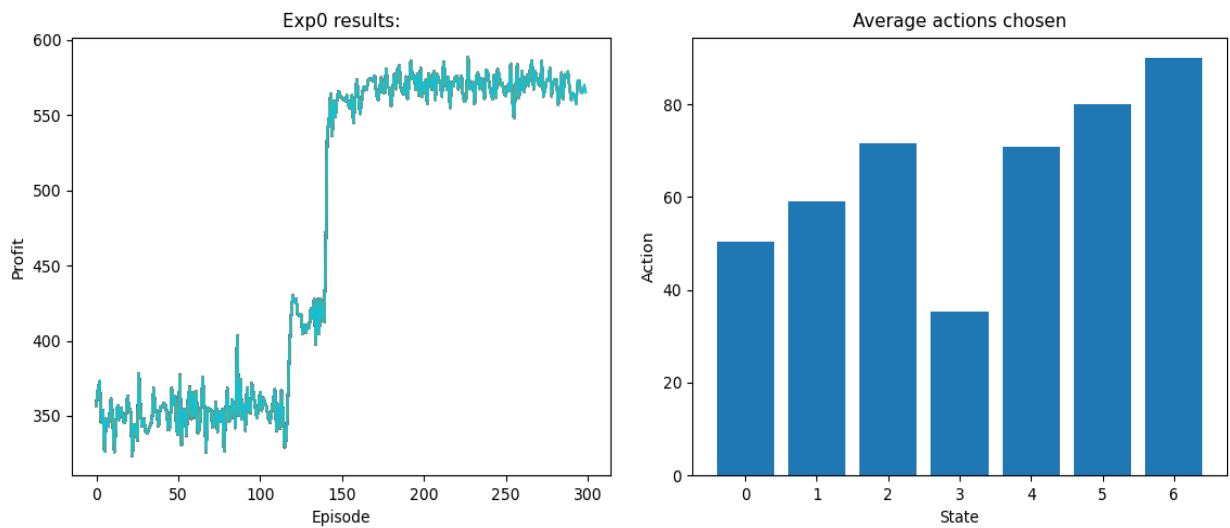

Figure 4: Plots from the main experiment

The learning curve has a similar shape to that of the first experiment, which is expected because the parameters remain the same. The right hand plot shows that the actions taken change dramatically across different states. Looking at days 0-2, the behavior here upholds our expectations from above, as the standard deviation correlates positively with the action. At day 3, the exponential distribution, the action is significantly lower, at around 35. This is a distribution with a single parameter 1/5, shifted over 30, so solving analytically yields:

$CDF = 1 - e^{-x/5} = \frac{15-5}{15} \;\; -> \;\; x = -5ln(\frac{1}{3}) = 5.49,$ so shifting back over 30 gives an optimum of 35.49. Again, this is very close, and demonstrates that the agent is able to learn a differently shaped, non-symmetrical distribution. Days 4-6 validate our sanity test to a reasonable degree, as the actual values are within 1-2% of their respective expectations. This experiment as a whole strongly achieves its goal of using the state space to help the agent concurrently learn several different distributions.

**4.3 Optimization**

There were a few miscellaneous measures I used to optimize the model's performance and reduce noise. For the output of the actor network, instead of just returning the fully connected sum, I restricted the range in the following way: The output of the layers is passed through a tanh function, which compresses all values within a range of (-1, 1). The actor is also passed minimum and maximum action values from the environment, and uses this to linearly scale up the value to a range of (min action, max action). This allows the actor to smoothly distribute its actions over a specific range, which helps the actor find the optimal actions efficiently.

I also made slight alterations to the internal reward structure to reduce noise. For the initial sanity tests, where the demand is the same every time, the agent was erroring significantly from the expectation. This indicated that the differences in rewards it was observing were too similar, and that at a certain point the positive rewards all 'blended together' to the agent. This prompted a modification where the reward returned from the environment is cubed before being stored in memory. This exaggerates the higher rewards even more, and scales everything in the *same direction*. This was able to reduce the noise to a negligible level and make the model robust.

### 4.4 Hyperparameters

The values of all hyperparameters used in the model are as follows:

| | | |
|---|---|---|
| τ | 0.005 | Rate at which to update the target networks |
| Batch Size | 256 | Number of transitions sampled in each training cycle |
| Noise STD ratio | 1/30 | Ratio of the action size to the amount of noise added to actions |
| Buffer Priming Period | 10000 | Timestep at which to begin training |
| Evaluation Timestep | 25000 | Timestep at which to begin evaluation |
| γ | 0.99 | Discount factor |
| LR | 1e-3 | Learning rate of each network |
| Actor net layers | 7x64x32x16x1 | The number of neurons in each layer, from input to output |
| Critic net layers | 8x64x32x16x1 | The number of neurons in each layer, from input to output |
| Buffer Size | 100000 | Maximum number of transitions to store |
| Episode length | 140 | Timesteps for each episode |
| # Episodes | 300 | Number of episodes in each experiment |

## 5. Conclusion

This paper has introduced a new and innovative way of solving the Newsvendor problem. I have shown that a DDPG reinforcement learning model is capable of learning how to maximize profit in a random demand environment. Additionally, the model is able to adapt to a wide variety of distributions, and learn their respective optimums simultaneously. With an environment design that is very flexible, this model can be molded to fit any inventory situation. This approach is completely data driven, and is more practical than purely mathematical models that rely on imperfect assumptions. RL here is able to improve upon traditional deep learning in its ability to adapt to real time data, and pick up on subtleties invisible to humans. I present this research to explore the ever-growing applications of RL and admire the power of our technology.